\documentclass[nohyperref]{article}

\usepackage{microtype}
\usepackage{graphicx}
\usepackage{subfigure}
\usepackage{booktabs} 
\usepackage{bbm}
\usepackage{hyperref}

\usepackage[accepted]{icml2022}

\usepackage{amsmath}
\usepackage{amssymb}
\usepackage{mathtools}
\usepackage{amsthm}

\usepackage[capitalize,noabbrev]{cleveref}

\theoremstyle{plain}
\newtheorem{theorem}{Theorem}[section]

\newtheorem{lemma}[theorem]{Lemma}

\theoremstyle{definition}

\theoremstyle{remark}

\usepackage[textsize=tiny]{todonotes}

\icmltitlerunning{Understanding Robust Generalization in Learning Regular Languages}

\begin{document}

\twocolumn[
\icmltitle{Understanding Robust Generalization in Learning Regular Languages}




\begin{icmlauthorlist}
\icmlauthor{Soham Dan}{yyy}
\icmlauthor{Osbert Bastani}{yyy}
\icmlauthor{Dan Roth}{yyy}
\end{icmlauthorlist}

\icmlaffiliation{yyy}{Department of Computer and Information Science, University of Pennsylvania}

\icmlcorrespondingauthor{Soham Dan}{sohamdan@seas.upenn.edu}
\icmlcorrespondingauthor{Osbert Bastani}{obastani@seas.upenn.edu}
\icmlcorrespondingauthor{Dan Roth}{danroth@seas.upenn.edu}
\icmlkeywords{Machine Learning, ICML}

\vskip 0.3in
]



\printAffiliationsAndNotice{}  

\begin{abstract}
A key feature of human intelligence is the ability to generalize beyond the training distribution, for instance, parsing longer sentences than seen in the past. Currently, deep neural networks struggle to generalize robustly to such shifts in the data distribution. We study robust generalization in the context of using recurrent neural networks (RNNs) to learn regular languages. We hypothesize that standard end-to-end modeling strategies cannot generalize well to systematic distribution shifts and propose a compositional strategy to address this. We compare an end-to-end strategy that maps strings to labels with a compositional strategy that predicts the structure of the deterministic finite state automaton (DFA) that accepts the regular language. We theoretically prove that the compositional strategy generalizes significantly better than the end-to-end strategy. In our experiments, we implement the compositional strategy via an auxiliary task where the goal is to predict the intermediate states visited by the DFA when parsing a string. Our empirical results support our hypothesis, showing that auxiliary tasks can enable robust generalization. Interestingly, the end-to-end RNN generalizes significantly better than the theoretical lower bound, suggesting that it is able to achieve at least some degree of robust generalization.
\end{abstract}

\section{Introduction}

\begin{figure}[!ht]
\centering
\includegraphics[width=0.45\textwidth]{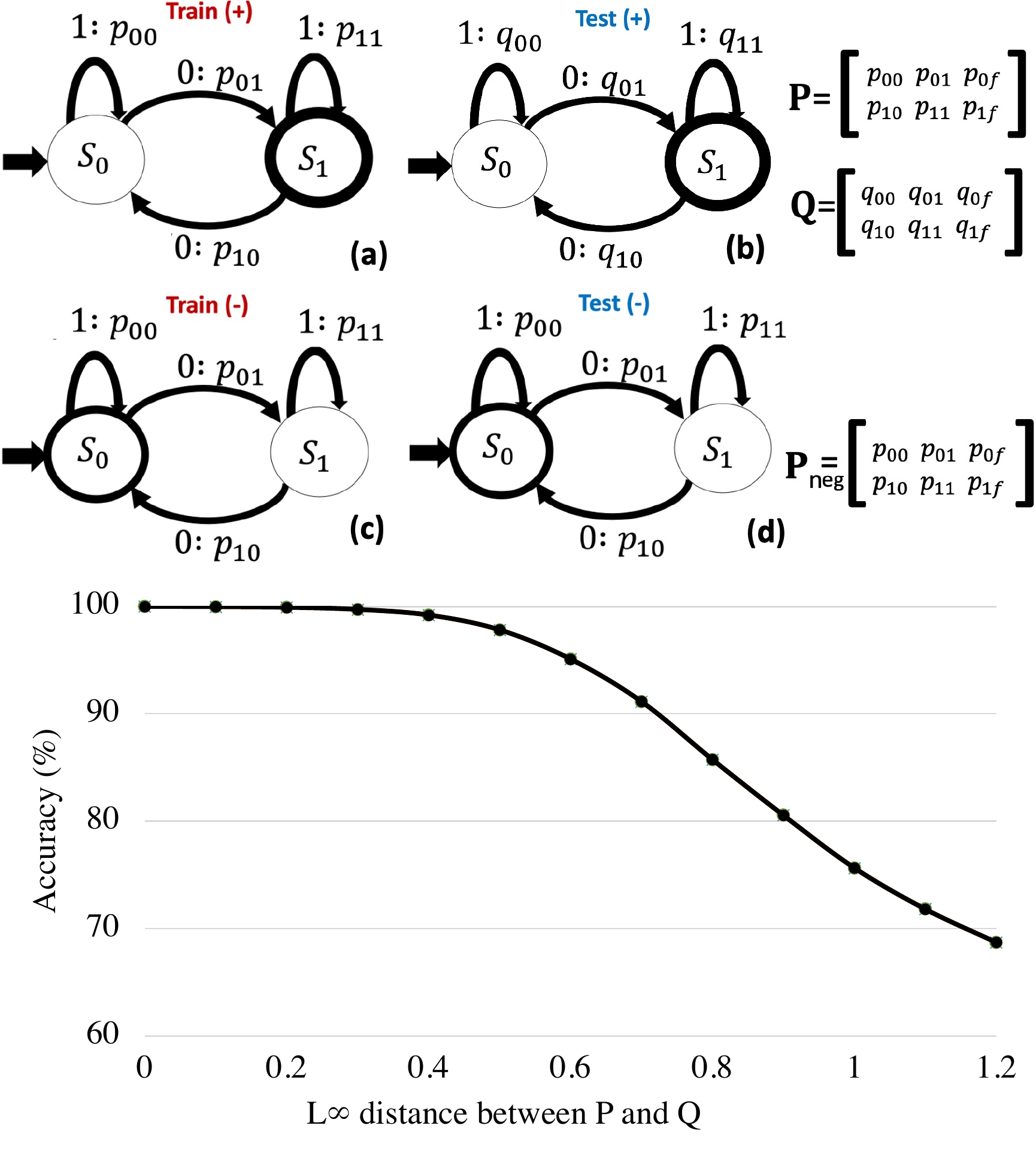}
\caption{The \emph{parity task} is to classify whether $x\in\{0,1\}^*$ has an odd ($y=1$) or even $(y=0)$ number of zeros. We show the edge Markov Chains for generating (a) training, and (b) shifted test examples. We also plot test accuracies as a function of $\|P-Q\|_\infty$, where $P$ and $Q$ are the transition probabilities of the edge Markov chains (the train accuracy is always $100\%$). The test accuracy drops significantly as $\|P-Q\|_\infty$ increases. We plot accuracy w.r.t the $\ell_\infty$ norm in order to compare with our theoretical bounds as discussed in Sec. \ref{subsec:bounds}.} 
\label{fig:intro}
\end{figure}

A key challenge facing deep learning is its inability to generalize \emph{robustly} to shifts in the underlying distribution. 
For example, \citet{lake2018generalization} demonstrate that recurrent neural networks (RNNs) have difficulty in generalizing to longer sentences than those seen during training and is also unable to understand novel combinations of familiar components, i.e., they lack \textit{systematic compositionality}.
Subsequent work have provided evidence of this failure across different architectures \cite{dessi2019cnns,furrer2020compositional} and tasks \cite{ruis2020benchmark,kim2020cogs}.

Humans, on the other hand, are remarkably robust to such shifts, suggesting that robust generalization is possible in practice. Thus, a key question is how to design deep learning algorithms that achieve this property. In the theoretical direction, most work has focused on learning in the presence of \emph{covariate shift}---i.e., shift in the distribution over inputs \cite{blitzer2008learning,pagnoni2018pac}. However, these results largely rely on the shift being small (in particular, bounded total variation (TV) distance), which does not hold for many of the shifts considered in robust generalization. 

Alternatively, the empirical work has focused primarily on devising new generalization splits on which existing models fail \cite{lake2018generalization, hupkes2020compositionality}. However,
without theoretical grounding, it is hard to interpret what degree of generalization can reasonably be expected.

In this paper, we consider the problem of classifying regular languages, which is sufficiently simple that we can theoretically analyze generalization in this setting. We consider two approaches to train neural networks. First, we consider an end-to-end approach that learns a model directly mapping strings to labels; this approach represents how neural networks are typically trained. Second, we consider a \emph{compositional} approach that is given access to intermediate supervision on the sequence of states visited by a deterministic finite state automaton (DFA) representing the language. Based on this extra information, the compositional strategy learns a model to predict the DFA transitions and final states.

We consider a training distribution in the form of a Markov model over strings that is constructed based on the DFA (essentially a Markov chain, but where emissions are on the edges); then, our goal is to generalize in the presence of shifts to this Markov model (Fig. \ref{fig:intro}). Small shifts in the Markov model probabilities can lead to large shifts in the distribution over strings---for instance, leading to significantly longer strings on average.

We provide two theoretical characterizations for each approach. First, we prove closed-form lower bounds on generalization of each strategy; these bounds rely on bounding the TV distance between the training and test distributions. Our results suggest that compositional approaches can generalize significantly better than the end-to-end strategy. Intuitively, compositionality helps since the distribution over DFA transitions shifts significantly less than the distribution over strings.

In practice, our bounds on the total variation distance can be overly conservative. Thus, we additionally propose algorithms for producing unbiased estimates of the relevant TV distances, and use them to empirically study generalization. In our experiments, the end-to-end strategy is a recurrent neural network (RNN) trained to directly map strings to labels. For the compositional strategy, we train an RNN in the same way, but provide it with an auxiliary task where on each step, its goal is to predict the current state of the underlying DFA. Intuitively, this strategy should align the RNN representation with the DFA state, making it easy to predict whether the DFA accepts a given string.

Our empirical results demonstrate that the compositional strategy generalizes significantly better than the end-to-end strategy. Interestingly, however, the end-to-end strategy generalizes significantly better than expected according to the estimated bound, suggesting that even end-to-end learning can exhibit some degree of robust generalization. Finally, our compositional strategy relies on additional supervision; we demonstrate that it is possible to achieve some degree of robust generalization even if we relax this supervision. 



\paragraph{Contributions.}
We formalize the problem of learning regular languages (Section~\ref{sec:prob}). In the context of this problem, we provide a theoretical analysis of robust generalization for both end-to-end and compositional learning, including theoretical bounds demonstrating the benefits of compositional learning, along with algorithms for obtaining unbiased estimates of the relevant TV distances (Section~\ref{sec:theory}). We then empirically demonstrate that an auxiliary task of predicting the DFA state achieves the compositional generalization bound and significantly outperforms the end-to-end strategy, though the end-to-end strategy exhibits some degree of robust generalization (Section~\ref{sec:exp}). We also perform experiments to analyze the impact of ``free'' auxiliary signals, the number of examples, model cell choice, length of examples, and the confidence calibration of the end-to-end and the compositional models on the shifted distribution. 


\section{Related Work}

\paragraph{Linguistics, automata, and RNNs.}
Linguistics has traditionally been tightly coupled to automata theory \cite{knight2009applications,suresh2021approximating}. Markov used finite state processes to predict sequences of vowels and consonants in novels \cite{markov1956essai,jurafskyspeech}, and Shannon extended this to predict letter sequences of English words using Markov processes \cite{shannon2001mathematical}. Markov chains and DFAs have applications in transliteration, translation, lexical processing, speech recognition, optical character recognition, summarization, sequence tagging, and in sequence generation such as speech synthesis and text generation. \cite{knight2009applications,gales2008application}. \par Recurrent neural networks (RNNs), which have been incredibly effective in natural language processing applications have renewed interest in automata theory in both linguistics and machine learning communities. RNNs have expressiveness closely connected to that of DFAs \cite{rabusseau2019connecting,michalenko2019representing,chen2018recurrent,tivno1998finite}. For instance, it is possible to extract the DFA from an RNN trained on sequences generated by that DFA \cite{weiss2018extracting,giles1992extracting,giles1992learning,omlin1996extraction,gers2001lstm,firoiu1998learning}; there has also been work trying to relate states of an RNN with those of a DFA \cite{tivno1998finite, michalenko2019representing}. These properties make RNNs ideal models for us to study. While existing work has focused on showing that RNNs can represent DFAs, their ability to learn DFAs in a way that generalize robustly has not yet been studied.


\paragraph{Systematic generalization in deep learning.}
The ability for neural networks to generalize robustly has been a longstanding question in cognitive science \cite{fodor1988connectionism}. Recently, several benchmarks have been proposed to investigate systematic generalization on carefully crafted train/test splits (e.g., the test set contains longer sequences than the training set) \cite{lake2018generalization,lake2019compositional,hupkes2020compositionality, loula2018rearranging,tsarkov2020cfq,ruis2020benchmark,kim2020cogs}. Although they are all motivated to measure systematic compositionality, there is no theoretical framework to understand the generalization for the different choices of splits, making it hard to know if generalization is at all possible for a given split; our goal in this paper is to take a step towards bridging this gap.

\paragraph{Learning theory and covariate shift.}
There has been significant theoretical work studying generalization in the presence of \emph{covariate shift} \cite{blitzer2008learning} (where the input distribution changes), with a large focus on \emph{domain adaptation} (i.e., where we are given unlabeled examples from the target domain). This has been widely studied in machine learning \cite{pagnoni2018pac,blitzer2008learning,zhang2019bridging,koh2021wilds} and natural language processing \cite{ben2021proceedings, li2012literature}. \citet{redko2020survey} surveys theoretical work in this area. Robust generalization is essentially covariate shift, enabling us to adapt techniques from this area.
One key challenge is that in general, learning with covariate shift is only possible if the shift is small (e.g., small TV distance), yet the shifts in the robust generalization settings we consider, are typically large. Thus, we must leverage additional structure to learn in a provably generalizable way; we prove that compositional learning can do exactly this by leveraging the DFA structure.

\section{Learning Regular Languages}
\label{sec:prob}

We formalize the learning problem we study. To do so, we need to define the following objects: (i) a classifier $f^*:\mathcal{X}\to\mathcal{Y}$ that we want to learn, (ii) a training distribution $P$ over $x\in\mathcal{X}$, and (iii) a test distribution $Q$ over inputs $x\in\mathcal{X}$. Then, the problem is to train a model $\hat{f}$ on inputs $x_1,...,x_n\sim P$ with labels $y_i=f^*(x_i)$, and then test $\hat{f}$ on inputs $x_1',...,x_{n'}'\sim Q$ with labels $y_i'=f^*(x_i')$.

The classifier $f^*$ is defined by a regular language $L(M)$ represented by a deterministic finite-state automaton (DFA) $M=(S,\Sigma,\delta,s_0,F)$, where $S$ is a finite set of states, $\Sigma$ is the alphabet, $\delta:S\times\Sigma\to S$ is the state transition function, $s_0\in S$ is the initial state, and $F\subseteq S$ is a set of final states~\cite{sipser1996introduction}. Given a string $x=\sigma_1...\sigma_T\in\Sigma^*$, we call $T$ the \emph{length} of $x$, and letting
\begin{align*}
s_t=\begin{cases}s_0&\text{if}~t=1\\\delta(s_{t-1},\sigma_{t-1})&\text{otherwise},\end{cases}
\end{align*}
we call $z=s_1...s_{T+1}\in S^*$ the state sequence \emph{induced} by $x$. We define $L(M)\subseteq \Sigma^*$ to be the strings accepted by $M$---i.e., $\sigma_1...\sigma_T\in L(M)$ if the induced state sequence $s_1...s_{T+1}$ satisfies $s_{T+1}\in F$. We let $\mathcal{X}=\Sigma^*$ be the strings over $\Sigma$, let $\mathcal{Y}=\{0,1\}$, and let $f^*(x)=\mathbbm{1}(x\in L(M))$ indicate whether $x$ is accepted by $M$.

Next, $P$ is defined by converting $M$ to a variant of a Markov chain. In particular, we assume given (i) probabilities $P(\sigma\mid s)$ of emitting $\sigma$ in state $s$, and (ii) probabilities $P(e\mid s)$ (where $e\in\{0,1\}$) of terminating upon reaching state $s$. Then, we sample $x\sim P$ as follows: initialize $s_1\gets s_0$; on each step $t$, terminate with probability $P(e\mid s)$ and return $x=\sigma_1...\sigma_t$; otherwise, sample $\sigma_t\sim P(\cdot\mid s_t)$, transition $s_{t+1}=\delta(s_t,\sigma_t)$, and continue. We call $P$ an \emph{edge Markov chain} since it is essentially a Markov chain with edge emissions.

This strategy only samples positive examples $x\in L(M)$; to sample negative examples, we use the same strategy with the automaton $M'$ for the complement $\mathcal{X}\setminus L(M)$. We sample positive and negative examples in equal proportion. Finally, we define the test distribution $Q$ similarly.

\section{Theoretical Analysis}
\label{sec:theory}

Next, we characterize the ability of algorithms for learning DFAs to generalize to shifted distributions, considering both the case where $\hat{f}$ directly maps sequences to labels, as well as a compositional strategy where $\hat{f}$ learns the DFA structure.

\subsection{Background on Covariate Shift}

We provide background on generalization bounds in the presence of \emph{covariate shift}---i.e., when the training and test distributions $P$ and $Q$ differ. Let
\begin{align*}
L_P(\hat{f})=\mathbb{P}_{x\sim P}[\hat{f}(x)\neq f^*(x)]
\end{align*}
be the loss of $\hat{f}$ on $P$, and similarly for $L_Q(\hat{f})$. Letting $\text{TV}(P(x),Q(x))=\sum_{x\in\mathcal{X}}|P(x)-Q(x)|$ be the total variation distance, we have the following.
\begin{lemma}
\label{lem:key}
$L_Q(\hat{f})\le L_P(\hat{f})+\text{TV}(P(x),Q(x))$
\end{lemma}
We give a proof in Appendix~\ref{sec:lem:key:proof}. In other words, we can characterize the loss of $\hat{f}$ on the test distribution $Q$ in terms of its loss on the training distribution $P$ and $\text{TV}(P(x),Q(x))$.

\subsection{Learning DFAs with Covariate Shift}

Next, we provide bounds on $\text{TV}(P(x),Q(x))$. We focus on a single pair of edge Markov chains $P(x)$ and $Q(x)$; given bound $\text{TV}(P^+(x),Q^+(x))\le\epsilon^+$ for the positive example distributions and bound $\text{TV}(P^-(x),Q^-(x))\le\epsilon^-$ for the negative example distributions, it is easy to check that
\begin{align*}
\text{TV}(P(x),Q(x))\le\frac{\epsilon^++\epsilon^-}{2},
\end{align*}
where $P(x)=(P^+(x)+P^-(x))/2$ and similarly for $Q(x)$. First, we have the following worst-case bound on the shift, specialized to the case where all strings are of a fixed length $T$.
\begin{lemma}
\label{lem:dfashift}
$\text{TV}(P(x),Q(x))\le2T|S|^{T+1}\epsilon$, where $\epsilon=\max_{s\in S}\text{TV}(P(\sigma\mid s),Q(\sigma\mid s))$
\end{lemma}
We give a proof in Appendix~\ref{sec:lem:dfashift:proof}. 
\begin{theorem}
\label{thm:dfashift}
$L_Q(\hat{f})\le L_P(\hat{f})+2T|S|^{T+1}\epsilon$
\end{theorem}
This result follows immediately from  Lemmas~\ref{lem:key} \&~\ref{lem:dfashift}. This result says that the accuracy of $\hat{f}$ decays exponentially in the length of the inputs, and it decays linearly in $\epsilon$, which quantifies the shift in the emission distributions of $P$ and $Q$.
However, this bound may be overly conservative. Given edge Markov chains $P(x)$ and $Q(x)$, we can estimate $\text{TV}(P(x),Q(x))$ using the following.
\begin{theorem}
\label{thm:dfashiftestimate}
We have
\begin{align*}
\text{TV}(P(x),Q(x))=\mathbb{E}_{x\sim P}\left[\left|1-\frac{Q(x)}{P(x)}\right|\right].
\end{align*}
\end{theorem}
We give a proof in Appendix~\ref{sec:thm:dfashiftestimate:proof}. Thus, given samples $x_1,...,x_n\sim P$, we have
\begin{align*}
\text{TV}(P(x),Q(x))\approx\sum_{i=1}^n\left|1-\frac{Q(x_i)}{P(x_i)}\right|.
\end{align*}
Given $x=\sigma_1...\sigma_T$, we can compute
\begin{align*}
P(x)=\prod_{t=1}^TP(\sigma_t\mid s_t),
\end{align*}
where $s_1...s_{T+1}$ is the state sequence induced by $x$, and similarly for $Q(x)$. Note that in this strategy, $T$ is not fixed and can vary with $x$.

\subsection{Compositional Learning of DFAs}

So far, we have considered a classifier trained directly to predict labels from strings. Next, we consider a compositional strategy that is given access to the hidden states of the DFA, learns to predict transitions and final states, and then composes these predictions to form the overall prediction.

Consider a model $\hat{g}:S\times\Sigma\to S$ trained to predict transitions (i.e., $\hat{g}(s,\sigma)\approx\delta(s,\sigma)$), along with a model $\hat{h}:S\to\{0,1\}$ trained to predict final states (i.e., $\hat{h}(s)\approx\mathbbm{1}(s\in F)$). Then, we have
\begin{align*}
\hat{f}(x)=\mathbbm{1}(s_{T+1}\in F),
\end{align*}
where $s_1...s_{T+1}$ is the state sequence induced by $x$. In this case, because $g$ and $h$ take states as inputs, we directly consider shifts in the state distribution. In particular, the distribution $P_t$ of states encountered by $\hat{g}$ and $\hat{h}$ on step $t$ is given by $P_t(s')=\mathbbm{1}(s'=s_0)$ if $t=1$, and
\begin{align*}
P_t(s')=\mathbb{P}_{s\sim P_{t-1},\sigma\sim P(\cdot\mid s)}[s'=\delta(s,\sigma)]
\end{align*}
otherwise, and similarly for $Q_t$. In addition, we let
\begin{align*}
P_t(s,\sigma)=P_t(s)P(\sigma\mid s).
\end{align*}
Then, $\hat{g}$ is trained on the distribution $P_t(s,\sigma)$ (for $t\in\{1,...,T\}$), and $\hat{h}$ is trained on $P_{T+1}(s)$. Similar to before, we have the following worst-case bound on the shift, specialized to the case where all strings are of a fixed length $T$.
\begin{lemma}
\label{lem:compshift}
We have $\text{TV}(P_t(s),Q_t(s))\le(t-1)\epsilon$, and $\text{TV}(P_t(s,\sigma),Q_t(s,\sigma))\le t\epsilon$.
\end{lemma}
We give a proof in Appendix~\ref{sec:lem:compshift:proof}.
\begin{theorem}
\label{thm:compshift}
$L_Q(\hat{f})\le\tilde{L}_P(\hat{f})+2T^2\epsilon$, where $\epsilon=\max_{s\in S}\text{TV}(P(\sigma\mid s),Q(\sigma\mid s))$ and
\begin{align*}
\tilde{L}_P(\hat{f})=\sum_{t=1}^TL_{P_t}(\hat{g})+L_{P_{T+1}}(\hat{h})
\end{align*}
\end{theorem}
We give a proof in Appendix~\ref{sec:thm:compshift:proof}. In this case, the accuracy of $\hat{f}$ scales quadratically in $T$, which is significantly better than Theorem~\ref{thm:dfashift}, and linearly $\epsilon$, which is the same as Theorem~\ref{thm:dfashift}. As before, this bound can be overly conservative, so we estimate the error for given edge Markov chains $P(x)$ and $Q(x)$ based on the following.
\begin{theorem}
\label{thm:compshiftestimate}
We have
\begin{align*}
L_Q(\hat{f})
&\le\tilde{L}_P(\hat{f})+\sum_{t=1}^T\text{TV}(P_t(s,\sigma),Q_t(s,\sigma)) \\
&\qquad\qquad\qquad+\text{TV}(P_{T+1}(s),Q_{T+1}(s))
\end{align*}
\end{theorem}
This result follows by the same argument as the proof of Theorem~\ref{thm:compshift}. We can estimate $P_t(s)$ by drawing samples $x_1,...,x_n\sim P$, and computing
\begin{align*}
\hat{P}_t(s)=\frac{1}{n}\sum_{i=1}^n\mathbbm{1}(s=s_{i,t}),
\end{align*}
where $s_{i,1}...s_{i,T_i+1}$ is the state sequence induced by $x_i$, and similarly for $\hat{Q}_t(s)$. Then, we have
\begin{align*}
\text{TV}(P_t(s),Q_t(s))\approx\sum_{s\in S}|\hat{P}_t(s)-\hat{Q}_t(s)|.
\end{align*}
Finally, we have $\hat{P}_t(s,\sigma)=\hat{P}_t(s)P(\sigma\mid s)$, and similarly for $\hat{Q}_t(s,\sigma)$, in which case
\begin{align*}
\text{TV}(P_t(s,\sigma),Q_t(s,\sigma))\approx\sum_{s\in S}|\hat{P}_t(s,\sigma)-\hat{Q}_t(s,\sigma)|.
\end{align*}
These estimates can be used in conjunction with Theorem~\ref{thm:compshiftestimate}. In our experiments, we make a modification---to reduce variance, rather than compute estimates for each $t$ separately, we aggregate states across all steps and use the average for all $\hat{P}_t(s)$, and similarly for $\hat{Q}_t(s)$. Finally, we heuristically take $T$ to be the average length of $x\sim P$.

\section{Experiments}
\label{sec:exp}

Next, we describe our experiments investigating whether end-to-end models can learn regular languages in a way that generalizes to distribution shifts, and whether different types of auxiliary supervision can help do so, and therefore enable robust generalization.

\subsection{Experimental Setup}

\begin{figure}
\centering
\includegraphics[width=0.48\textwidth]{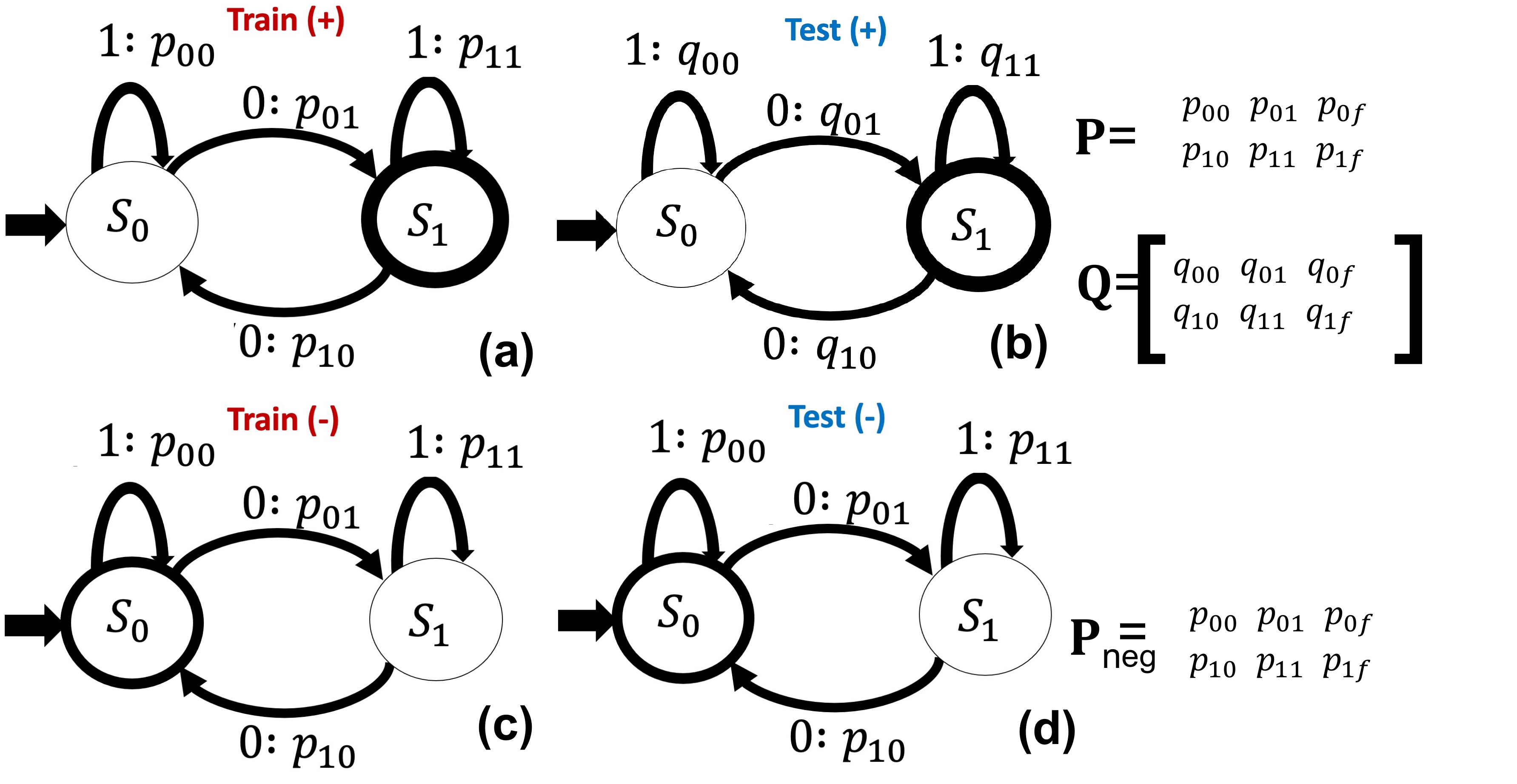}
\caption{The eMC\textsubscript{id} and eMC\textsubscript{ood} used to generate the train and o.o.d. test positive examples, and the eMC\textsubscript{neg} used to generate negative examples. Here, $\rightarrow$ denotes the start state; bold circles denote end states; $P$, $Q$, and $P_{neg}$ denote the transition matrices for eMC\textsubscript{id}, eMC\textsubscript{ood}, and eMC\textsubscript{neg}, respectively; and $0f,1f$ denotes the end probability in states $S_0$ and $S_1$, respectively, with $p_{0f}=0$ in eMC\textsubscript{id} and $p_{1f}=0$ in eMC\textsubscript{ood}.}
\label{fig:wfst}
\end{figure}

\paragraph{Classification problem.}
We consider the regular language classification problem.
We construct an edge Markov chain eMC\textsubscript{id}
to generate training examples and in-domain (i.d.) test examples for each language by assigning transition probabilities to each edge of the DFA for that language. To generate out-of-domain (o.o.d.) test examples, we perturb some of the edge probabilities of eMC\textsubscript{id} to obtain eMC\textsubscript{ood}. To generate negative examples, we use the same strategy for the complement language $\mathcal{L}^C=\mathcal{X}\setminus\mathcal{L}$
to obtain eMC\textsubscript{neg}; for simplicity, we do not shift the negative example distribution.

We focus on the parity language $\mathcal{L}_{par}$ as a case-study, and include additional results for other languages in Appendix \ref{appendix:additionl_exp}. This language has $\Sigma=\{0,1\}$, and consists of all strings containing an odd number of zeros.
Our DFA for $\mathcal{L}_{par}$ has two states $S=\{s_0,s_1\}$, where $s_0$ is the start state, and final states $F=\{s_1\}$.
Fig. \ref{fig:wfst} shows the edge-Markov chains
eMC\textsubscript{id}, eMC\textsubscript{ood}, and eMC\textsubscript{neg} we use. Note that by increasing the loop probabilities, we generate longer sequences; thus, the o.o.d. test distribution contains longer strings on average than the training distribution.

We also consider a broader class of \textit{modulo languages}: regular languages over $\Sigma = \{ 0,1 \}$ of the form $\mathcal{L}_{mod-k}$, where the number of zeros are a multiple of $k$,
for $k\in\{3,4,5\}$.

\paragraph{Classifier.} We train an RNN binary classifier on examples generated by eMC\textsubscript{id} (positive) and eMC\textsubscript{neg} (negative). We evaluate it on i.d. test examples generated by eMC\textsubscript{id}, eMC\textsubscript{neg}, and on o.o.d. test examples from eMC\textsubscript{ood},eMC\textsubscript{neg}.

To improve performance, we propose \emph{state sequence auxiliary supervision (SSAS)}, where we additionally train the RNN on an auxiliary supervised learning task. In particular, given an input example $x\sim P$, where $P$ is an edge Markov chain, in addition to training the RNN to predict the ground truth label $f^*(x)$, we train its representation $z_t$ at each step $t$ to predict the state $s_t$ visited by $P$ at step $t$. This supervision task is a multi-class classification problem, so we use the cross-entropy loss. This auxiliary loss is jointly optimized with the binary classification loss for the main task (the losses are weighted equally). Importantly, the auxiliary supervision signal is only provided during training; thus, at test time, the RNN acts identically to an RNN trained only on the main task (i.e., without SSAS).

We use an RNN with LSTM cells, with an embedding dimension of $50$ and a hidden layer with dimension $50$, optimized using stochastic gradient descent (SGD) with a learning rate of $0.01$ \footnote{Hyper-parameters chosen based on accuracy on i.d. dev-set.}. We use $N_{train}^+=1600$ positive and $N_{train}^-=1600$ negative train examples, $N_{dev}^+= N_{dev}^-=200$ dev examples, and use $N_{test}^+=2000$ positive and $N_{test}^-=2000$ negative examples for each of the i.d. and o.o.d. test sets.


\begin{table}[]
\centering
\begin{tabular}{lrrr} 
\toprule
\multicolumn{1}{c}{\textbf{Task}} &
\multicolumn{1}{c}{\textbf{E2E}} &
\multicolumn{1}{c}{\textbf{SSAS}} &
\multicolumn{1}{c}{\textbf{RI} (\%)} \\
\midrule
$\mathcal{L}_{mod-3}$
& 66.10 & \textbf{98.68} & 49.49 \\ 
$\mathcal{L}_{mod-4}$
& 64.61 & \textbf{97.58}  & 51.03 \\ 
$\mathcal{L}_{mod-5}$
& 65.15 & \textbf{93.00} & 42.75 \\ 
\bottomrule
\end{tabular}
\caption{Accuracy (\%) of the end-to-end (E2E) vs. SSAS training on the o.o.d. test set, with the
relative improvements (RI) of SSAS over E2E for the \textit{modulo languages}: $\mathcal{L}_{mod-3}$, $\mathcal{L}_{mod-4}$, $\mathcal{L}_{mod-5}$. }
\label{tab:my_label}
\end{table}

\paragraph{Metrics.}
We perform several different experiments to compare our theoretical bounds with the empirical accuracy, studying the impact of the number of training examples, model cell variations (vanilla RNN, LSTM, or GRU), asymmetry in length generalization, the use of ``free'' auxiliary tasks that do not require knowledge of the DFA, and the confidence calibration of the models.


\subsection{End-to-End vs. SSAS Training}

First, we compare the o.o.d. accuracy of end-to-end and SSAS training (in both cases, the i.d. test accuracy is $100\%$). Table \ref{tab:my_label} shows the o.o.d. test accuracy of each approach for the case $\text{TV}(P,Q)=1.3$ for each of the modulo-languages considered \footnote{Additional plots for the \textit{modulo languages} are in Appendix \ref{appendix:additionl_exp}.}. As can be seen, SSAS training significantly improves accuracy compared to end-to-end training. Strictly speaking, SSAS is not a compositional learning algorithm, but it guides the RNN to learn representations that correctly encode the compositional structure of the DFA. Thus, these results validate our theoretical finding that compositional training significantly improves generalization.

\subsection{Theoretical vs. Empirical Generalization}
\label{subsec:bounds}

\begin{figure}
\centering
\includegraphics[width=0.48\textwidth]{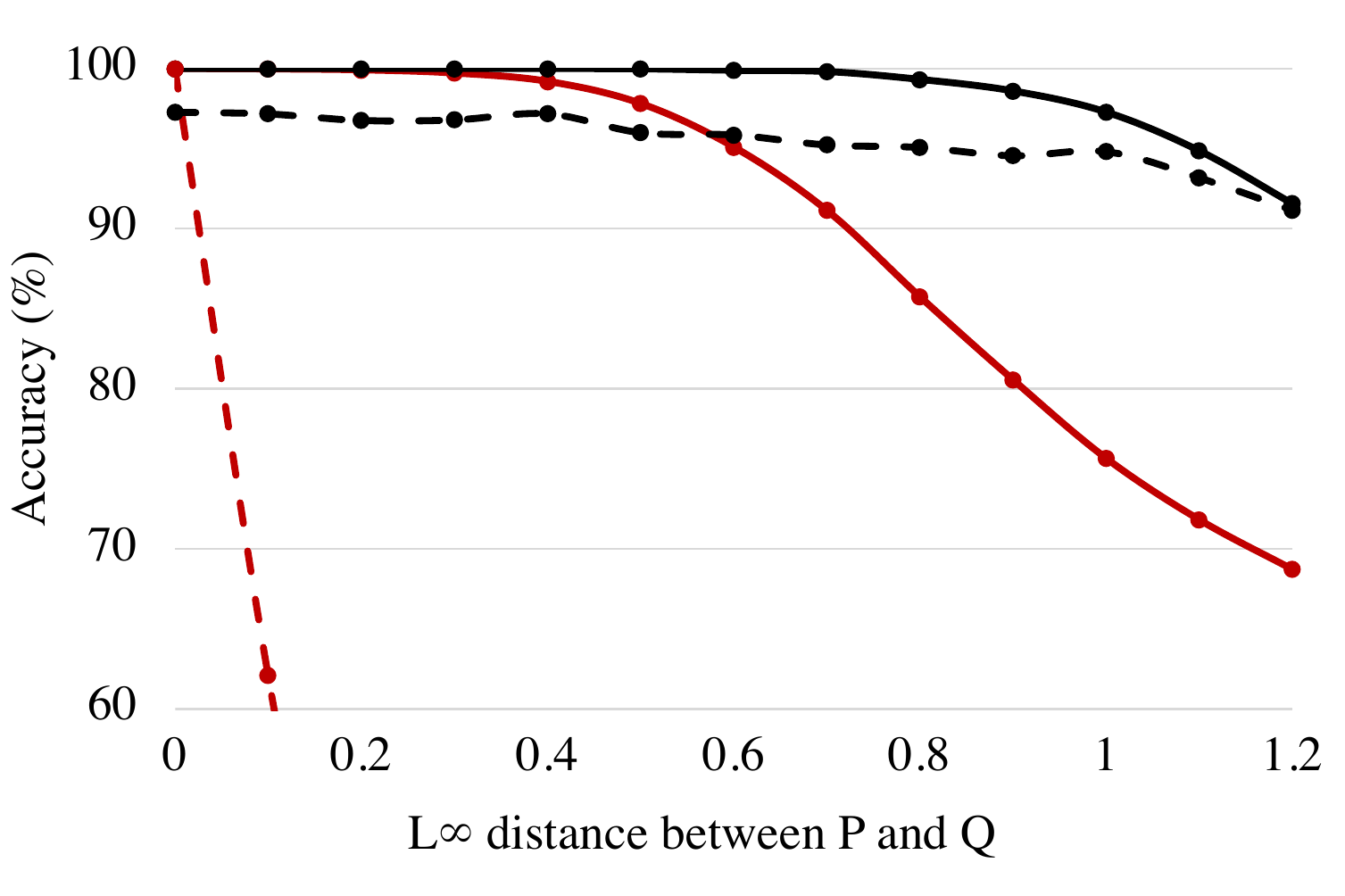}
\caption{We plot the empirical o.o.d. test set accuracies for the end-to-end model (red solid) and the SSAS model (black solid), and the theoretical estimate of the o.o.d. accuracies for the end-to-end model (red dashed) and the SSAS model (black dashed).}
\label{fig:bounds}
\end{figure}

Focusing on the parity language, we compare the empirical accuracy on the o.o.d. test set to our estimates of the accuracy based on Theorems~\ref{thm:dfashiftestimate} \&~\ref{thm:compshiftestimate} in Section~\ref{sec:theory}. 
We consider eMC\textsubscript{id} with transitions
\begin{align*}
P=\begin{bmatrix}
0.2 & 0.8 & 0 \\
0.7 & 0.2 & 0.1
\end{bmatrix},
\end{align*}
eMC\textsubscript{ood} with transitions
\begin{align*}
Q=\begin{bmatrix}
\delta & 1-\delta & 0 \\
0.9-\delta & \delta & 0.1
\end{bmatrix},
\end{align*}
where $\delta\in\{0.2,0.25,...,0.85\}$, and eMC\textsubscript{neg} with
\begin{align*}
P_{neg}=\begin{bmatrix}
0.7 & 0.2 & 1 \\
0.8 & 0.2 & 0
\end{bmatrix}.
\end{align*}
In Fig.~\ref{fig:bounds}, we plot the following four quantities as a function of $\|P-Q\|_\infty$ (for our choices of $P$ and $Q$, we have $\epsilon=2\|P-Q\|_\infty$, where $\epsilon$ is defined in Theorem~\ref{thm:dfashift}):
\begin{itemize}
\item Empirical accuracy on the o.o.d. test set (solid lines), for end-to-end (red) and SSAS (black)
\item Theoretical estimates of the accuracy (dashed lines) for end-to-end (red) and SSAS (black)
\end{itemize}
For the theoretical estimate of the accuracy: (i) for end-to-end, we draw $n=10000$ samples from $eMC_{id}$ and compute the lower bound based on Theorem~\ref{thm:dfashiftestimate}, and (ii) for SSAS, we use $10000$ samples from each $eMC_{id}$ and $eMC_{ood}$, and estimate the lower bound based on Theorem~\ref{thm:compshiftestimate}; for both, we average across $10$ random repetitions. Further, the reported empirical accuracies are the average over $10$ random runs.

As can be seen, the theoretical estimates are tight for SASS, indicating the SASS matches the expected generalization rate for compositional models. These results support our intuition that SASS enables the RNN to learn the compositional structure of the DFA. Interestingly, the end-to-end model significantly outperforms the theoretical estimate. Since the estimate of the TV distance converges to its true value, so the gap between the theoretical and empirical values must either be due to the inequality in Lemma~\ref{lem:key} or the fact that the RNN is learning some compositional structure. We expect that the latter must be happening to some degree to explain such a large gap---importantly, the gap is substantially larger than the gap for SASS.


\subsection{Free Auxiliary Signals}
\label{subsec:free_aux}

While SSAS improves generalization, a critical shortcoming is that it requires knowledge of the underlying DFA; in particular, it uses the DFA to construct ground truth state sequences used to train the RNN. In this section, we study whether auxiliary tasks computed without such prior knowledge can improve generalization. As an example, we consider the \textit{count auxiliary task}; intuitively, a model that can count zeros is more likely to correctly learn the parity concept. More precisely, we consider a $10$-class classification task, where zero counts greater than $8$ are represented by the $10^{th}$ class. As with SSAS, we equally weigh the cross-entropy loss for this auxiliary task with the binary cross-entropy loss for the main task.

\begin{figure}
\centering
\includegraphics[width=0.48\textwidth]{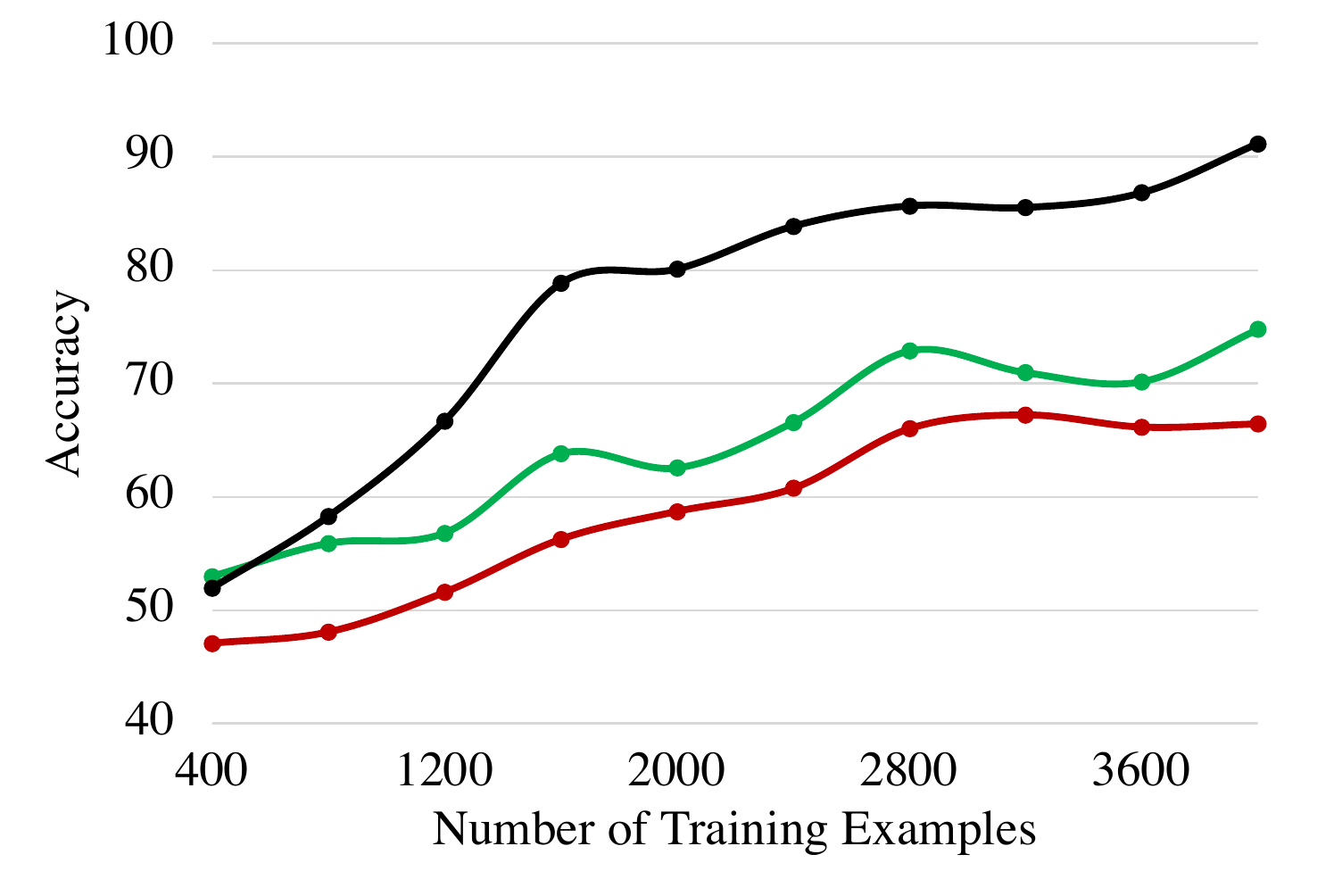}
\caption{For the setting described in Sec. \ref{subsec:free_aux}, we plot the accuracies of the baseline (red), SSAS model (black), and the model with free auxiliary supervision (green), for varying number of training examples. We see that \textit{auxiliary count supervision} helps the baseline model consistently across varying amounts of training data. }
\label{free_aux}
\end{figure}

We let $P$, $Q$, and $P_{neg}$ be as in Section~\ref{subsec:bounds}, with $\delta=0.85$.
In Fig.~\ref{free_aux}, we plot the o.o.d. test accuracy (estimated on $4000$ examples) as a function of the number of training examples, ranging from $400$ to $4000$. As can be seen, the count auxiliary task improves performance by $5-8\%$. The improvement is less than in SSAS, which is to be expected since the task in SSAS gives information directly relevant to the main task. These results demonstrate that even without knowledge of the DFA, we can improve generalization via auxiliary supervision.


\subsection{Effect of Number of Training Examples}
\label{number}

Next, we show that simply increasing the number of training examples is insufficient for achieving robust generalization, 
demonstrating that novel techniques such as SSAS are required to generalize robustly. 
\par
\begin{figure}
\centering
\includegraphics[width=0.48\textwidth]{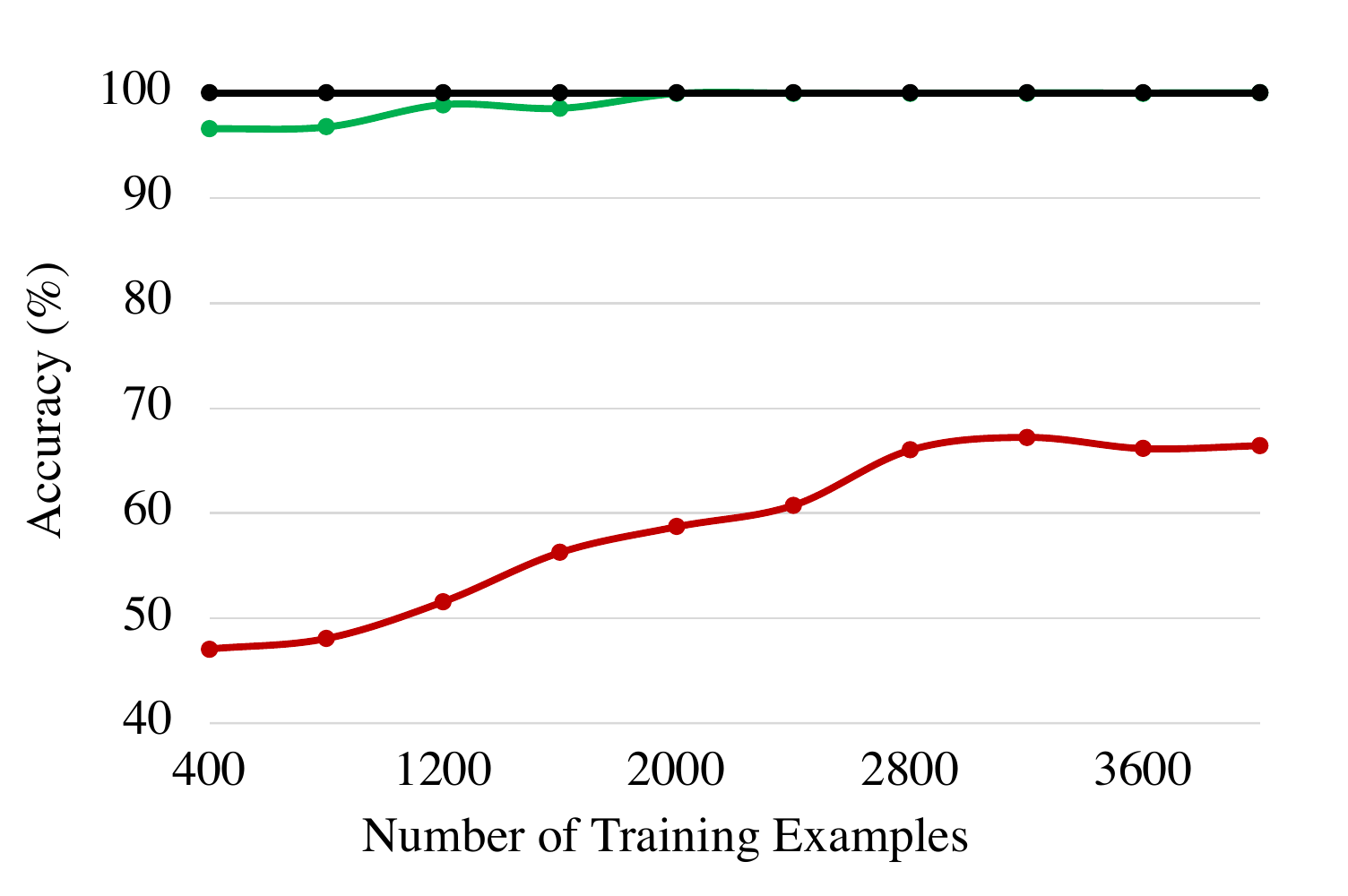}
\caption{We plot the train (black), i.d. test (green), and o.o.d. test (red) accuracies as a function of the number of training examples drawn from the $eMC_{id}$ for the baseline (end-to-end) model.}
\label{fig:num_bounds}
\end{figure}

 We consider $P$, $Q$, and $P_{neg}$ as in Section~\ref{subsec:bounds}, with $\delta=0.85$.
In Fig.~\ref{fig:num_bounds}, we plot the empirical train, i.d. test, and o.o.d. test accuracies as a function of the the number of training examples varying from $400$ to $4000$; in all cases, we estimate accuracy on $4000$ test examples. As can be seen, the i.d. test accuracy quickly reaches $100\%$, but the o.o.d. test accuracy
remains significantly lower, stabilizing around $66\%$ at $2800$ training examples (the train accuracy is $100\%$ throughout). These results also justify our choice of $3200$ training examples in our other experiments.

\subsection{Effect of the Model Cell}
\label{subsec:cell}

Next, we study whether changes in the model cell architecture affect generalization. We consider long short term memory (LSTM) units, recurrent neural network (RNN) units, and gated recurrence units (GRU).
\par
\begin{figure}
\centering
\includegraphics[width=0.48\textwidth]{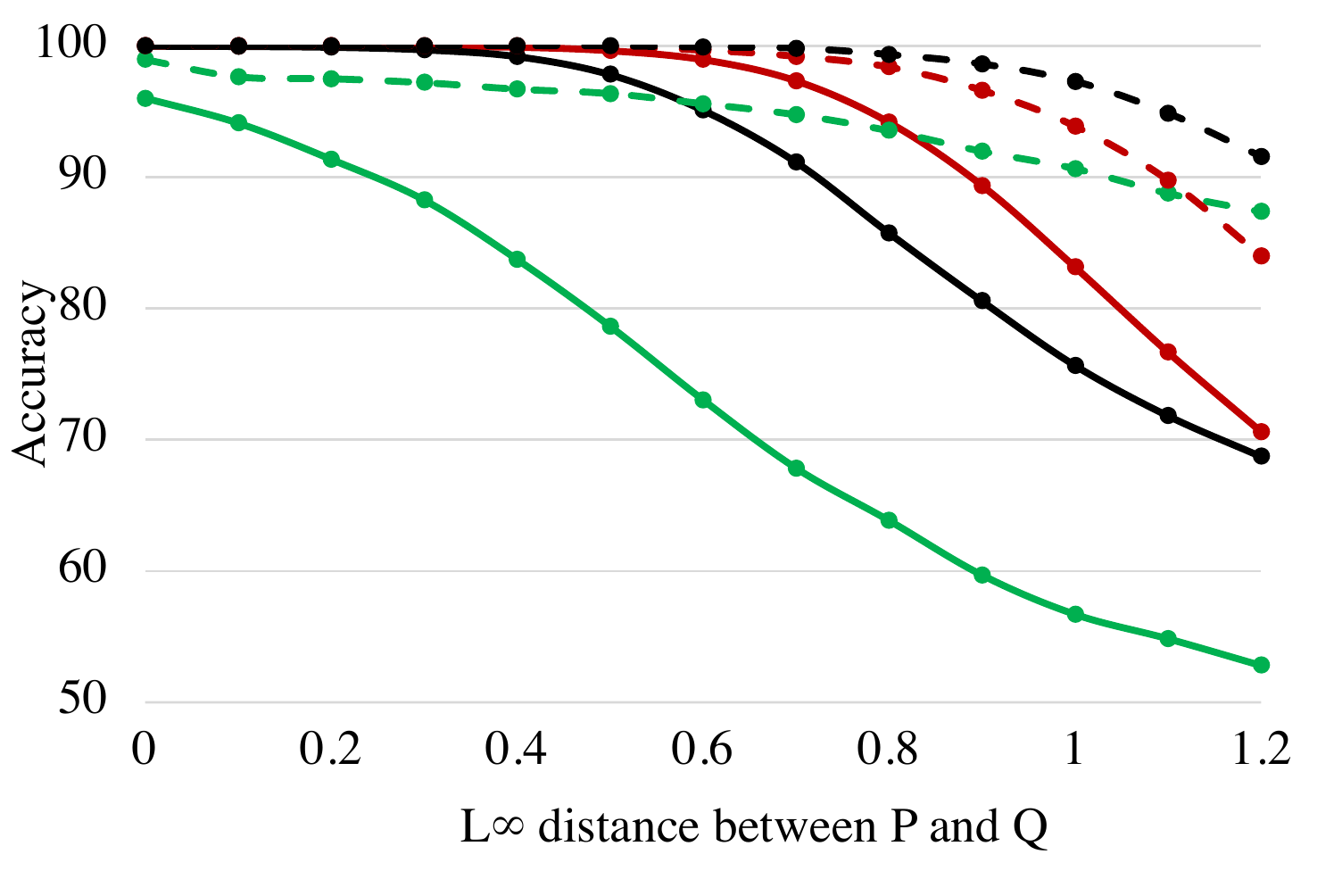}
\caption{We plot the accuracy of models trained end-to-end (solid) and using SSAS (dashed) as a function $\|P-Q\|_\infty$, for models with LSTM (black), GRU (red), and RNN (green) cells.}
\label{fig:cell}
\end{figure}

 We set $P$, $Q$, and $P_{neg}$ as in Section~\ref{subsec:bounds}. In Fig.~\ref{fig:cell}, we plot the o.o.d. test accuracy (estimated using $4000$ examples) as a function of $\|P-Q\|_\infty$ for each of the three choices. As before, the train and i.d. test accuracy are 100$\%$ for all choices. For models trained end-to-end, LSTM and GRU cells perform comparably in terms of o.o.d. accuracy whereas RNN cells perform significantly worse. These results are in line with the fact that RNNs are worse at capturing long-range dependencies, which are necessary for solving the parity task. For models trained using SSAS, LSTMs perform the best; interestingly, GRUs perform well at first but become worse than RNNs as $\|P-Q\|_\infty$ becomes large. The superior performance of LSTMs over other cell choices, justifies our use of them in the other experiments.

\subsection{Asymmetric Length Generalization}
\label{subsec:asymmetry}

\begin{figure}
\centering
\includegraphics[width=0.4\textwidth]{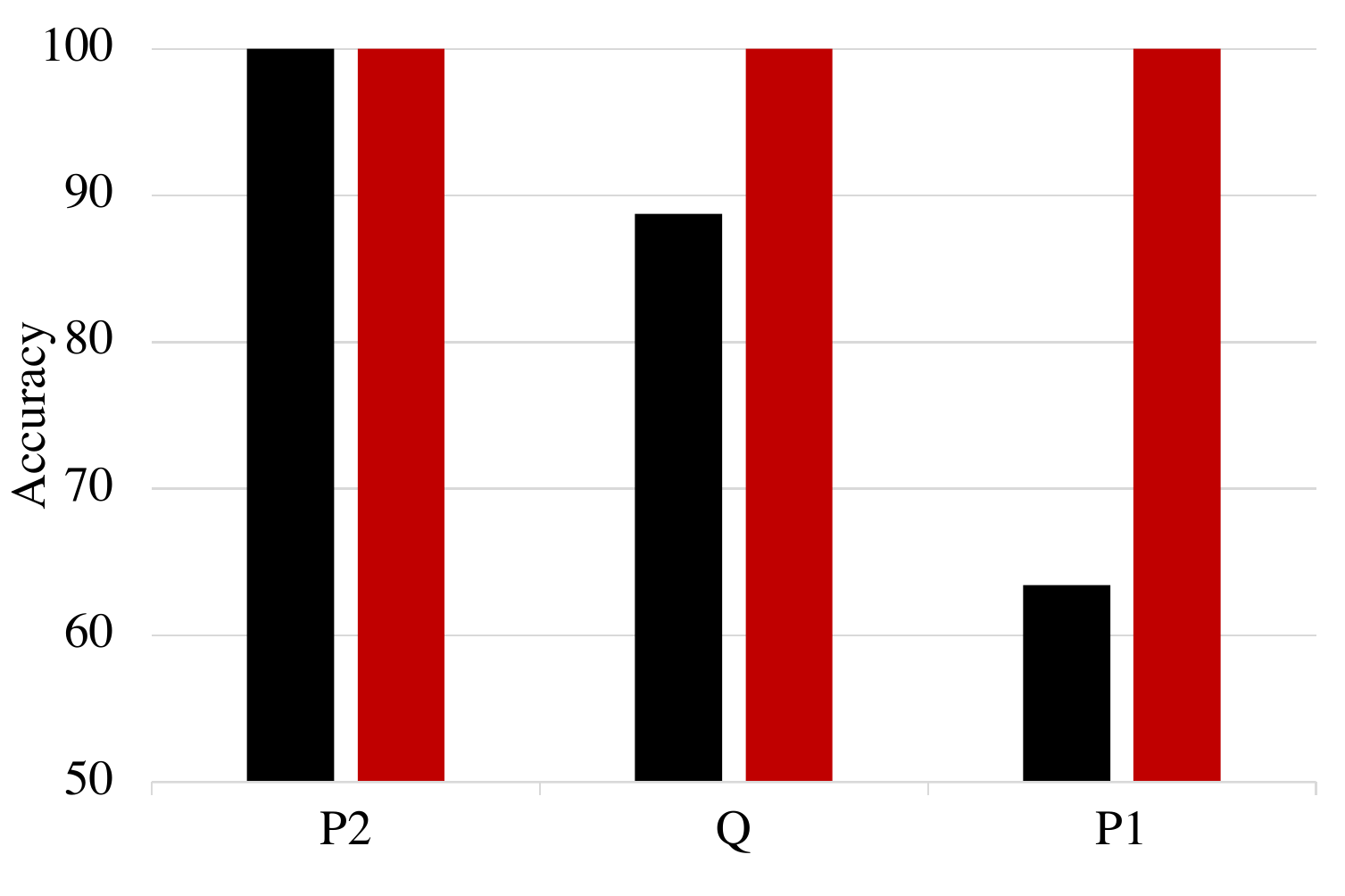}
\caption{As described in Section~\ref{subsec:asymmetry}, we train models on $3200$ examples generated from two different eMCs with transition matrices $P_1$ (red) and $P_2$ (black),
and tested on $4000$ test examples separately generated from three different eMCs $P_1$, $P_2$, and $Q$.}
\label{fig:asymmetry}
\end{figure}

Next, we study how different training distributions affects generalization. In particular, in the context of the parity language, we consider two training distributions given by edge Markov chains with transition probabilities $P_1$ and $P_2$ and a test distribution given by an edge Markov chain with transition probabilities $Q$, where
\begin{align*}
P_1&=\begin{bmatrix}
0.2 & 0.8 & 0 \\
0.7 & 0.2 & 0.1
\end{bmatrix} \\
P_2&=\begin{bmatrix}
0.8 & 0.2 & 0 \\
0.1 & 0.8 & 0.1
\end{bmatrix} \\
Q&=\begin{bmatrix}
0.5 & 0.5 & 0 \\
0.4 & 0.5 & 0.1
\end{bmatrix}.
\end{align*}
The negative examples are generated from eMC\textsubscript{neg} described in Section~\ref{subsec:bounds}. Importantly, we have $\|P_1-Q\|_\infty = \|P_2-Q\|_\infty=0.6$.
In Fig. \ref{fig:asymmetry}, we show the results of evaluating models trained on examples from $P_1$ (red) and $P_2$ (black) on test examples generated using $P_1$, $P_2$, and $Q$. As can be seen, the model trained on examples from $P_1$ generalizes well to test examples from $Q$ and $P_2$, whereas the model trained from $P_2$ generalizes poorly to test examples from $Q$ and $P_1$. In this case, $P_1$ tends to generate longer sequences than $P_2$ (with $Q$ being in between); thus, our results show that training on longer sequences can generalize to shorter sequences, whereas training on shorter sequences cannot generalize to training on longer sequences.

\subsection{Confidence Calibration}
\label{subsec:calibration}
We study confidence calibration of the baseline (E2E) and SSAS models as a function of increasing separation, $\|P-Q\|_\infty$. Calibration measures how well the posterior probabilities of a model are aligned with the empirical likelihoods \cite{guo2017calibration}. We use the following metrics: 
\par
    $\bullet$ \textit{Brier Score (BS)} \cite{brier1950verification} is a proper scoring rule for measuring the accuracy of predicted probabilities. It is defined as the mean squared error between the predicted probabilities and the actual targets. If $n$ denotes the total number of examples and $\Psi_i$ denotes the prediction probability that $f(x_i)=1$, then the Brier Score is:
    $$
    BS=\frac{1}{n}\sum_{i=1}^n (\Psi_i-y_i)^2.
    $$
    $\bullet$ \textit{Expected Calibration Error (ECE)} \cite{guo2017calibration} measures the difference in expectation between confidence and accuracy. Empirically this is approximated by dividing the data into $M$ confidence based bins, $B_1,...,B_M$, where $B_m$ contains all datapoints $x_i$ for which $\Psi_i$ lies in $(\frac{m-1}{M},\frac{m}{M}]$. If $acc(B_m)$ and $conf(B_m)$ denotes the average accuracy and prediction confidence for the points in $B_m$, then ECE is:
    $$
ECE = \sum_{m=1}^M \frac{|B_m|}{n} |acc(B_m) -conf(B_m)|.
$$
In Table \ref{tab:calibration} we see that while calibration degrades with increasing $\|P-Q\|_\infty$ for both models, SSAS is significantly better calibrated than E2E under the distribution shift. 

\begin{table}[]
    \centering
    \begin{tabular}{|c|c|c|c|c|c|}
    \hline
     Metric & Model & $0.0$ & $0.4$ & $0.8$ & $1.2$\\
     \hline
     BS & E2E & $5$e-$06$ & $9$e-$04$ & $0.073$ & $0.262$\\ \cline{2-6} 
     & SSAS &  \textbf{3e-06} & \textbf{1e-04} & $\textbf{0.008}$ & $\textbf{0.051}$ \\ 
     \hline \hline
     ECE & E2E & $4$e-$04$ & $0.003$ & $0.097$ & $0.303$\\ \cline{2-6} 
      & SSAS & \textbf{3e-04} & $\textbf{0.002}$ & $\textbf{0.035}$ & $\textbf{0.125}$ \\ \hline

\end{tabular}
    \caption{Brier Score (BS) and Expected Calibration Error (ECE) as a function of increasing $\|P-Q\|_\infty = \lbrace 0, 0.4, 0.8, 1.2 \rbrace$ for the E2E and SSAS models. Note that lower values are better.}
    \label{tab:calibration}
\end{table}

\section{Conclusion}

We have studied robust generalization of RNNs learning regular languages, providing both theoretical and empirical evidence that compositional strategies generalize more robustly to shifted distributions compared to end-to-end strategies. In particular, leveraging an auxiliary task in the form of state supervision (which we call SSAS) achieves the theoretical rate anticipated by our compositional generalization guarantee. We have also demonstrated that generic auxiliary tasks such as counting zeros can also improve generalization. Interestingly, we find that end-to-end learning outperforms the theoretical rate for end-to-end learning, suggesting that end-to-end approaches can still achieve some degree of robust generalization.

A key direction for future work is to explore the effectiveness of other techniques for enabling robust generalization. It would also be interesting to explore how these results generalize to other kinds of tasks beyond regular languages, as well as to study the performance of state-of-the-art architectures such as transformers in this setting.

\bibliography{example_paper}
\bibliographystyle{icml2022}

\newpage
\appendix
\onecolumn

\section{Proofs}

\subsection{Proof of Lemma~\ref{lem:key}}
\label{sec:lem:key:proof}

We have
\begin{align*}
L_Q(\hat{f})
=\sum_{x\in\mathcal{X}}\mathbbm{1}(\hat{f}(x) \neq f^*(x))Q(x)
&=\sum_{x\in\mathcal{X}}\mathbbm{1}(\hat{f}(x) \neq f^*(x))(P(x)+Q(x)-P(x)) \\
&\le L_P(\hat{f})+\sum_{x\in\mathcal{X}}|Q(x)-P(x)|,
\end{align*}
as claimed. $\qed$

\subsection{Proof of Lemma~\ref{lem:dfashift}}
\label{sec:lem:dfashift:proof}

The given emission probabilities $P(\sigma\mid s)$ induce transition probabilities over the DFA states---in particular, the probability of transitioning from $s$ to $s'$ is
\begin{align*}
P(s'\mid s)=\sum_{\sigma\in\Sigma}P(\sigma\mid s)\cdot\mathbbm{1}(s'=\delta(s,\sigma)).
\end{align*}
Then, the probability of a sequence of states is $P(s_1...s_{T'})=\prod_{t=1}^TP(s_{t+1}\mid s_t)$, where $T'=T+1$, and similarly for $Q(s_1...s_{T'})$. Now, we prove the claim. First, note that
\begin{align*}
|P(x)-Q(x)|
&=\left|\sum_{z\in S^{T'}}P(x\mid z)P(z)-\sum_{z\in S^{T'}}Q(x\mid z)Q(z)\right| \\
&\le\left|\sum_{z\in S^{T'}}(P(x\mid z)P(z)-P(x\mid z)Q(z))\right|+\left|\sum_{z\in S^{T'}}(P(x\mid z)Q(z)-Q(x\mid z)Q(z))\right| \\
&\le\sum_{z\in S^{T'}}P(x\mid z)|P(z)-Q(z)|+\sum_{z\in S^{T'}}|P(x\mid z)-Q(x\mid z)|Q(z).
\end{align*}
Thus, we have
\begin{align}
\text{TV}(P(x),Q(x))
&=\sum_{x\in\Sigma^T}|P(x)-Q(x)| \nonumber \\
&\le\sum_{z\in S^{T'}}\sum_{x\in\Sigma^T}P(x\mid z)|P(z)-Q(z)|+\sum_{z\in S^{T'}}\sum_{x\in\Sigma^T}|P(x\mid z)-Q(x\mid z)|Q(z) \nonumber \\
&=\text{TV}(P(z),Q(z))+\mathbb{E}_{z\sim Q}[\text{TV}(P(x\mid z),Q(x\mid z)]. \label{eqn:thm1:proof}
\end{align}
Consider the first term of (\ref{eqn:thm1:proof}). Letting $z=s_1...s_{T'}$, note that
\begin{align*}
|P(z)-Q(z)|
&=\left|\prod_{t=1}^TP(s_{t+1}\mid s_t)-\prod_{t=1}^TQ(s_{t+1}\mid s_t)\right| \\
&=\left|\sum_{\tau=1}^T\left(\prod_{t=1}^\tau P(s_{t+1}\mid s_t)\prod_{t=\tau+1}^TQ(s_{t+1}\mid s_t)-\prod_{t=1}^{\tau-1}P(s_{t+1}\mid s_t)\prod_{t=\tau}^TQ(s_{t+1}\mid s_t)\right)\right| \\
&=\left|\sum_{\tau=1}^T\left[\prod_{t=1}^{\tau-1}P(s_{t+1}\mid s_t)\right]\left[\prod_{t=\tau+1}^TQ(s_{t+1}\mid s_t)\right](P(s_{\tau+1}\mid s_t)-Q(s_{\tau+1}\mid s_t))\right| \\
&\le\sum_{\tau=1}^T|P(s_{\tau+1}\mid s_t)-Q(s_{\tau+1}\mid s_t))| \\
&\le T\max_{s\in S}\text{TV}(P(s'\mid s),Q(s'\mid s)).
\end{align*}
Furthermore, note that for any $s\in S$, we have
\begin{align}
\text{TV}(P(s'\mid s),Q(s'\mid s))
&=\sum_{s'\in S}|P(s'\mid s)-Q(s'\mid s)| \nonumber \\
&=\sum_{s'\in S}\left|\sum_{\sigma\in\Sigma}(P(\sigma\mid s)-Q(\sigma\mid s))\cdot\mathbbm{1}(s'=\delta(s,\sigma))\right| \nonumber \\
&\le\sum_{\sigma\in\Sigma}|P(\sigma\mid s)-Q(\sigma\mid s)|\sum_{s'\in S}\mathbbm{1}(s'=\delta(s,\sigma)) \nonumber \\
&=\text{TV}(P(\sigma\mid s),Q(\sigma\mid s)), \label{eqn:tvbound}
\end{align}
As a consequence, we have
\begin{align*}
\sum_{z\in S^{T'}}|P(z)-Q(z)|
&\le T|S|^{T'}\max_{s\in S}\text{TV}(P(s'\mid s),Q(s'\mid s)) \\
&\le T|S|^{T'}\max_{s\in S}\text{TV}(P(\sigma\mid s),Q(\sigma\mid s)).
\end{align*}
Now, consider the second term of (\ref{eqn:thm1:proof}). Letting $x=\sigma_1...\sigma_T$ and $z=s_1...s_{T'}$, we have
\begin{align*}
|P(x\mid z)-Q(x\mid z)|
&=\left|\prod_{t=1}^TP(\sigma_t\mid s_t)-\prod_{t=1}^TQ(\sigma_t\mid s_t)\right| \\
&=\left|\sum_{\tau=1}^T\prod_{t=1}^\tau P(\sigma_t\mid s_t)\prod_{t=\tau+1}^TQ(\sigma_t\mid s_t)-\prod_{t=1}^{\tau-1}P(\sigma_t\mid s_t)\prod_{t=\tau}^TQ(\sigma_t\mid s_t)\right| \\
&=\left|\sum_{\tau=1}^T\left[\prod_{t=1}^{\tau-1}P(\sigma_t\mid s_t)\right]\left[\prod_{t=\tau+1}^TQ(\sigma_t\mid s_t)\right](P(\sigma_\tau\mid s_\tau)-Q(\sigma_\tau\mid s_\tau))\right| \\
&\le\sum_{\tau=1}^T\left[\prod_{t=1}^{\tau-1}P(\sigma_t\mid s_t)\right]\left[\prod_{t=\tau+1}^TQ(\sigma_t\mid s_t)\right]|P(\sigma_\tau\mid s_\tau)-Q(\sigma_\tau\mid s_\tau))|.
\end{align*}
As a consequence, we have
\begin{align*}
&\sum_{x\in\Sigma^T}|P(x\mid z)-Q(x\mid z)| \\
&\le\sum_{\sigma_1...\sigma_T\in\Sigma^T}\sum_{\tau=1}^T\left[\prod_{t=1}^{\tau-1}P(\sigma_t\mid s_t)\right]\left[\prod_{t=\tau+1}^TQ(\sigma_t\mid s_t)\right]|P(\sigma_\tau\mid s_\tau)-Q(\sigma_\tau\mid s_\tau))| \\
&=\sum_{\tau=1}^T\left[\prod_{t=1}^{\tau-1}\sum_{\sigma_t\in\Sigma}P(\sigma_t\mid s_t)\right]\left[\prod_{t=\tau+1}^T\sum_{\sigma_t\in\Sigma}Q(\sigma_t\mid s_t)\right]\sum_{\sigma_\tau\in\Sigma}|P(\sigma_\tau\mid s_\tau)-Q(\sigma_\tau\mid s_\tau))| \\
&=\sum_{\tau=1}^T\sum_{\sigma_\tau\in\Sigma}|P(\sigma_\tau\mid s_\tau)-Q(\sigma_\tau\mid s_\tau))| \\
&=\sum_{\tau=1}^T\text{TV}(P(\sigma\mid s_\tau),Q(\sigma\mid s_\tau)).
\end{align*}
Therefore, we have
\begin{align*}
\mathbb{E}_{z\sim Q}[\text{TV}(P(x\mid z),Q(x\mid z)]
&=\sum_{\tau=1}^T\mathbb{E}_{z\sim Q}[\text{TV}(P(\sigma\mid s_\tau),Q(\sigma\mid s_\tau))] \\
&\le T\max_{s\in S}\text{TV}(P(\sigma\mid s),Q(\sigma\mid s)).
\end{align*}
The claim follows since $T\epsilon+T|S|^{T+1}\epsilon\le2T|S|^{T+1}\epsilon$. $\qed$

\subsection{Proof of Theorem~\ref{thm:dfashiftestimate}}
\label{sec:thm:dfashiftestimate:proof}

Note that
\begin{align*}
\text{TV}(P(x),Q(x))
=\sum_{x\in\mathcal{X}}|P(x)-Q(x)|
=\sum_{x\in\mathcal{X}}\left|1-\frac{Q(x)}{P(x)}\right|P(x)
=\mathbb{E}_{x\sim P}\left[\left|1-\frac{Q(x)}{P(x)}\right|\right],
\end{align*}
as claimed. $\qed$

\subsection{Proof of Lemma~\ref{lem:compshift}}
\label{sec:lem:compshift:proof}

We prove the first claim by induction on $t$. The base case $t=1$ is trivial, since $P_1(s)=Q_1(s)=\mathbbm{1}(s=s_0)$. For the inductive case, note that
\begin{align*}
&\text{TV}(P_t(s'),Q_t(s')) \\
&=\sum_{s'\in S}|P_t(s')-Q_t(s')| \\
&=\left|\sum_{s'\in S}\sum_{s\in S}P_{t-1}(s)P(s'\mid s)-Q_{t-1}(s)Q(s'\mid s)\right| \\
&\le\left|\sum_{s'\in S}\sum_{s\in S}P_{t-1}(s)P(s'\mid s)-P_{t-1}(s)Q(s'\mid s)\right|+\left|\sum_{s'\in S}\sum_{s\in S}P_{t-1}(s)Q(s'\mid s)-Q_{t-1}(s)Q(s'\mid s)\right| \\
&\le\sum_{s'\in S}\sum_{s\in S}P_{t-1}(s)|P(s'\mid s)-Q(s'\mid s)|+\sum_{s'\in S}\sum_{s\in S}|P_{t-1}(s)-Q_{t-1}(s)|Q(s'\mid s) \\
&\le\max_{s\in S}\text{TV}(P(s'\mid s),Q(s'\mid s))+\text{TV}(P_{t-1}(s),Q_{t-1}(s)) \\
&\le\max_{s\in S}\text{TV}(P(\sigma\mid s),Q(\sigma\mid s))+\text{TV}(P_{t-1}(s),Q_{t-1}(s)) \\
&\le(t-1)\epsilon,
\end{align*}
where the second-to-last step follows from (\ref{eqn:tvbound}) in the proof of Theorem~\ref{thm:dfashift}, so the first claim follows. For the second claim, note that
\begin{align*}
\text{TV}(P_t(s,\sigma),Q_t(s,\sigma))
&=\sum_{s\in S}\sum_{\sigma\in\Sigma}|P_t(s,\sigma)-Q_t(s,\sigma)| \\
&=\sum_{s\in S}\sum_{\sigma\in\Sigma}|P_t(s)P(\sigma\mid s)-Q_t(s)Q(\sigma\mid s)| \\
&\le\sum_{s\in S}\sum_{\sigma\in\Sigma}|P_t(s)P(\sigma\mid s)-P_t(s)Q(\sigma\mid s)|+|P_t(s)Q(\sigma\mid s)-Q_t(s)Q(\sigma\mid s)| \\
&=\sum_{s\in S}\sum_{\sigma\in\Sigma}P_t(s)|P(\sigma\mid s)-Q(\sigma\mid s)|+|P_t(s)-Q_t(s)|Q(\sigma\mid s) \\
&=\max_{s\in S}\text{TV}(P(\sigma\mid s),Q(\sigma\mid s))+\text{TV}(P_t(s),Q_t(s)) \\
&\le t\epsilon,
\end{align*}
so the second claim follows as well. $\qed$

\subsection{Proof of Theorem~\ref{thm:compshift}}
\label{sec:thm:compshift:proof}

First, by a union bound, we have
\begin{align*}
L_Q(\hat{f})
&=\mathbb{P}_{x\sim Q}[\hat{f}(x)\neq f^*(x)] \\
&=\mathbb{P}_{x\sim Q}\left[\bigvee_{t=1}^T\hat{g}(s_t,\sigma_t)\neq\delta(s_t,\sigma_t)\vee\hat{h}(s_{T+1})\neq\mathbbm{1}(s_{T+1}\in F)\right] \\
&\le\sum_{t=1}^T\mathbb{P}_{x\sim Q}[\hat{g}(s_t,\sigma_t)\neq\delta(s_t,\sigma_t)]+\mathbb{P}_{x\sim Q}[\hat{h}(s_{T+1})\neq\mathbbm{1}(s_{T+1}\in F)] \\
&=\sum_{t=1}^T\mathbb{P}_{(s_t,\sigma_t)\sim Q_t}[\hat{g}(s_t,\sigma_t)\neq\delta(s_t,\sigma_t)]+\mathbb{P}_{s_{T+1}\sim Q_{T+1}}[\hat{h}(s_{T+1})\neq\mathbbm{1}(s_{T+1}\in F)] \\
&=\sum_{t=1}^TL_{Q_t}(\hat{g})+L_{Q_{T+1}}(\hat{h}).
\end{align*}
By Lemmas~\ref{lem:key} \&~\ref{lem:compshift}, the loss of $\hat{g}$ on step $t$ satisfies
\begin{align*}
L_{Q_t}(\hat{g})\le L_{P_t}(\hat{g})+T\epsilon,
\end{align*}
and similarly the loss of $\hat{h}$ satisfies
\begin{align*}
L_{Q_{T+1}}(\hat{h})\le L_{P_{T+1}}(\hat{h})+T\epsilon.
\end{align*}
Thus, we have
\begin{align*}
L_Q(\hat{f})
&\le\left[\sum_{t=1}^TL_{P_t}(\hat{g})+L_{P_{T+1}}(\hat{h})\right]+T(T+1)\epsilon,
\end{align*}
so the claim follows since $T(T+1)\epsilon\le2T^2\epsilon$ (since $T\ge1$). $\qed$

\section{Additional Experiments}
\label{appendix:additionl_exp}
\begin{figure}[!b]
\centering
\includegraphics[width=0.3\textwidth]{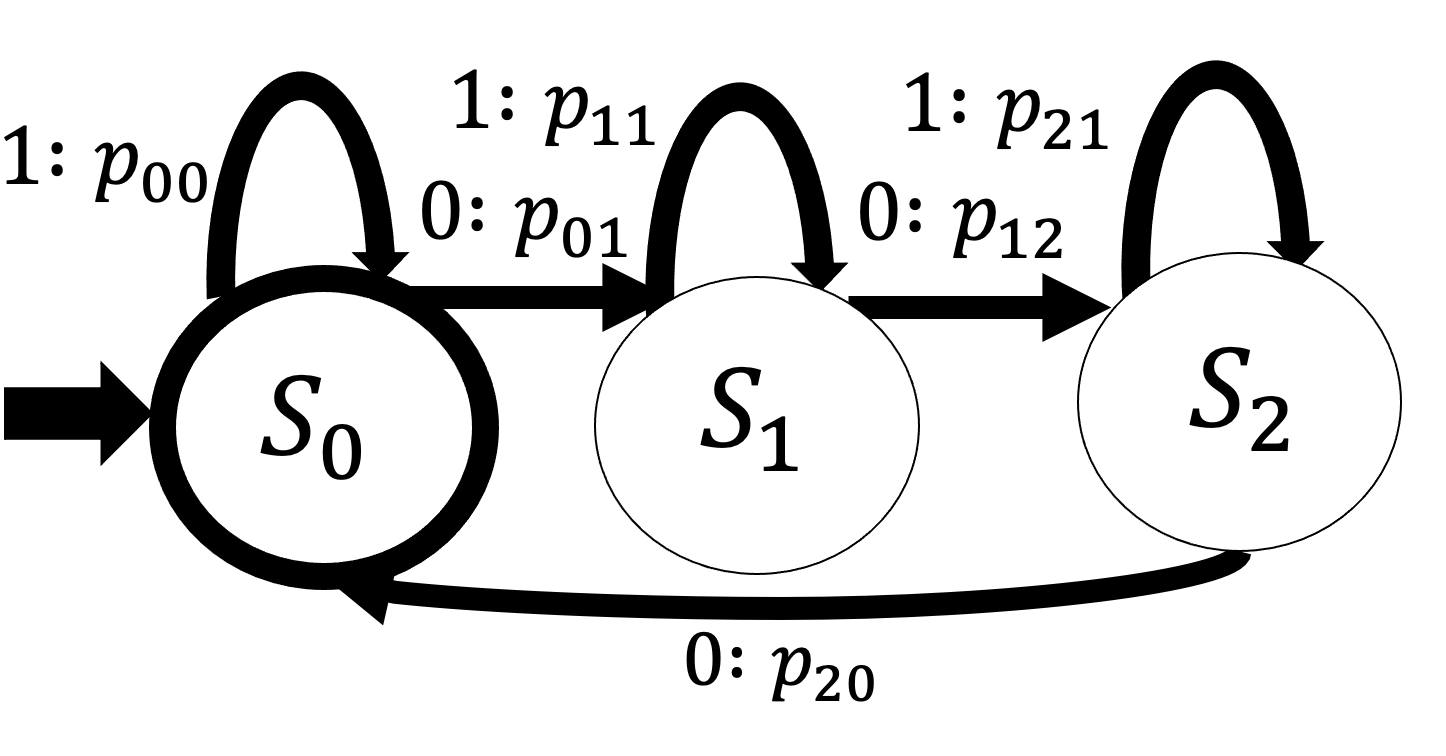} \\
\includegraphics[width=0.3\textwidth]{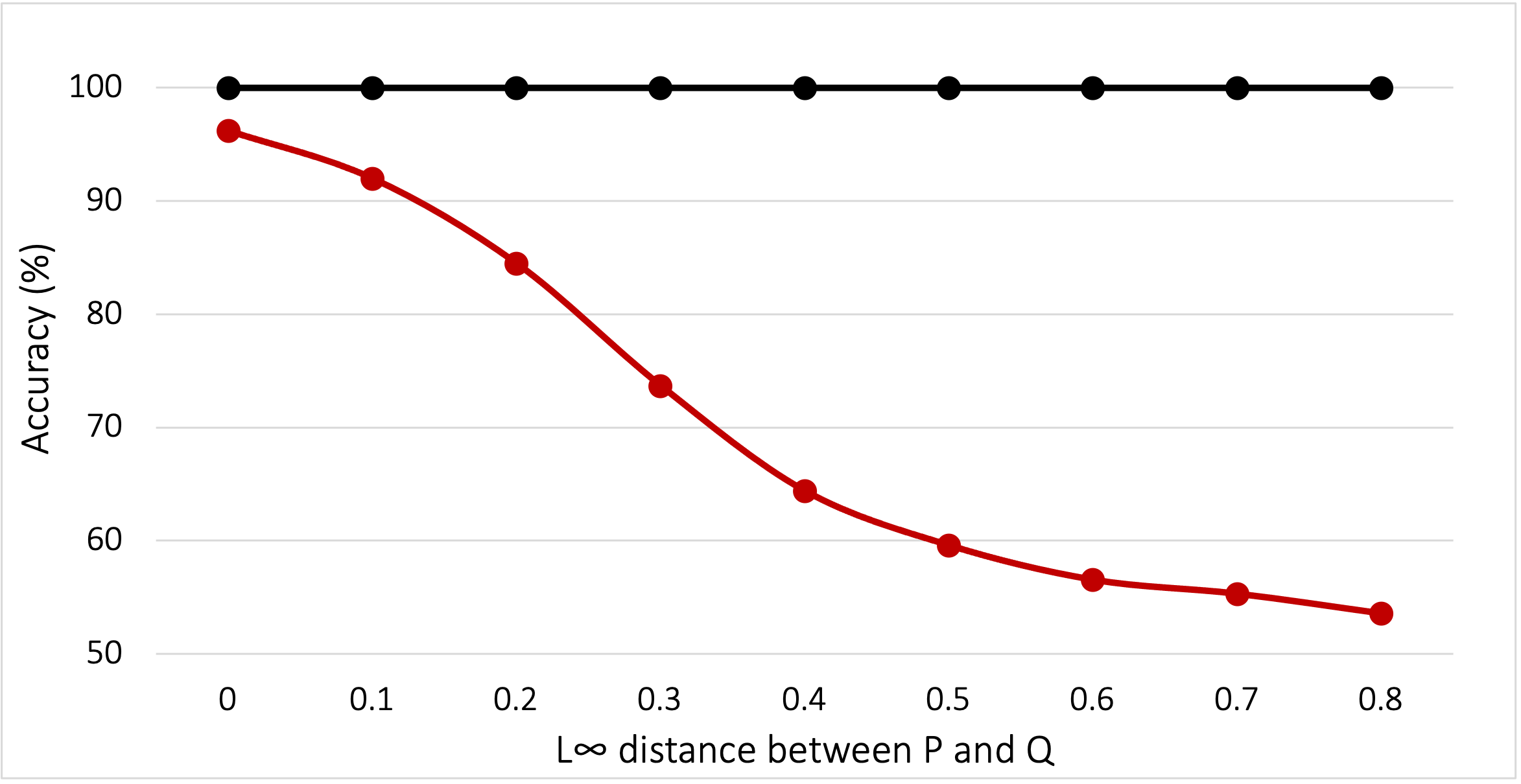}
\includegraphics[width=0.3\textwidth]{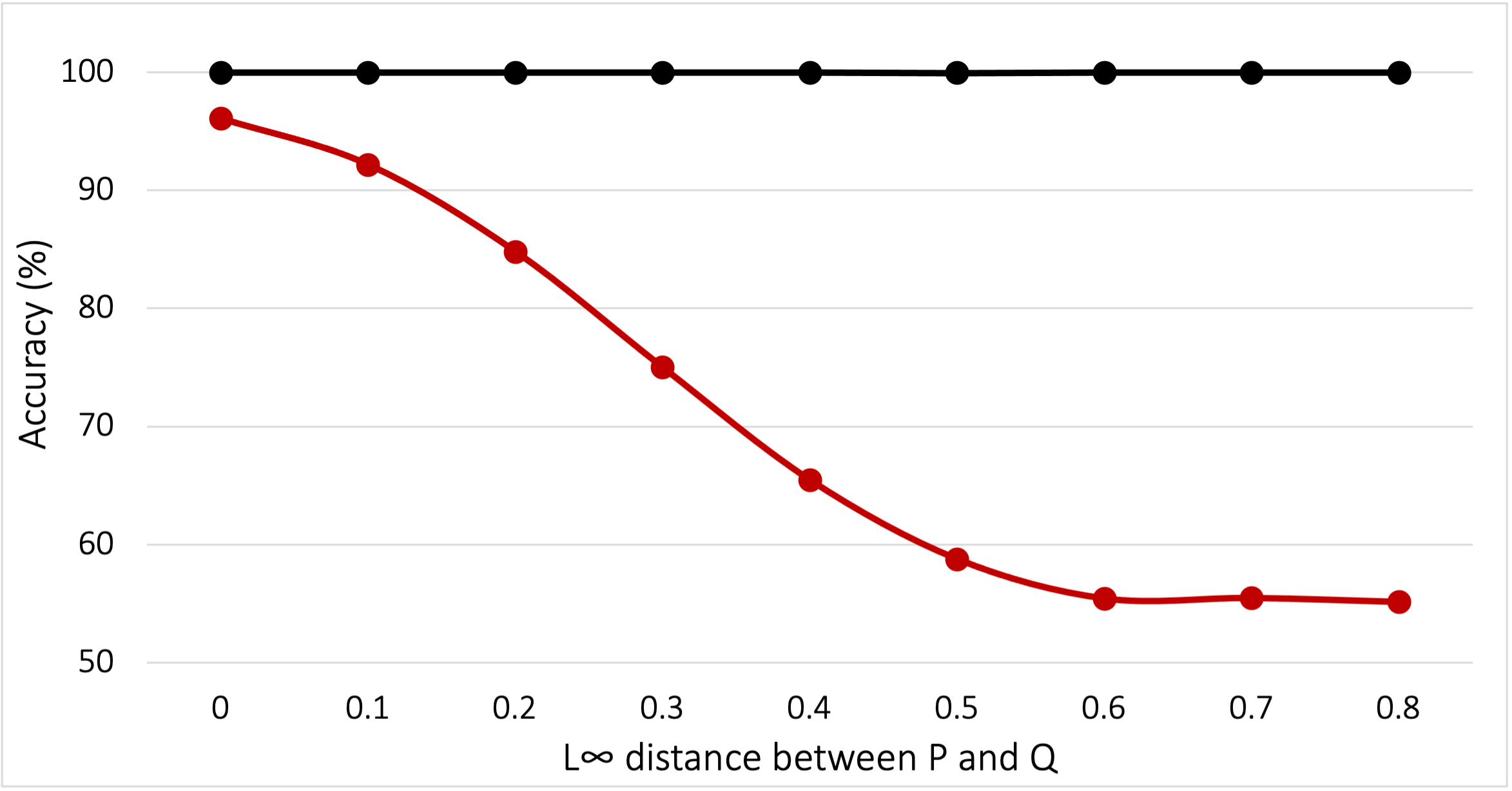}
\includegraphics[width=0.3\textwidth]{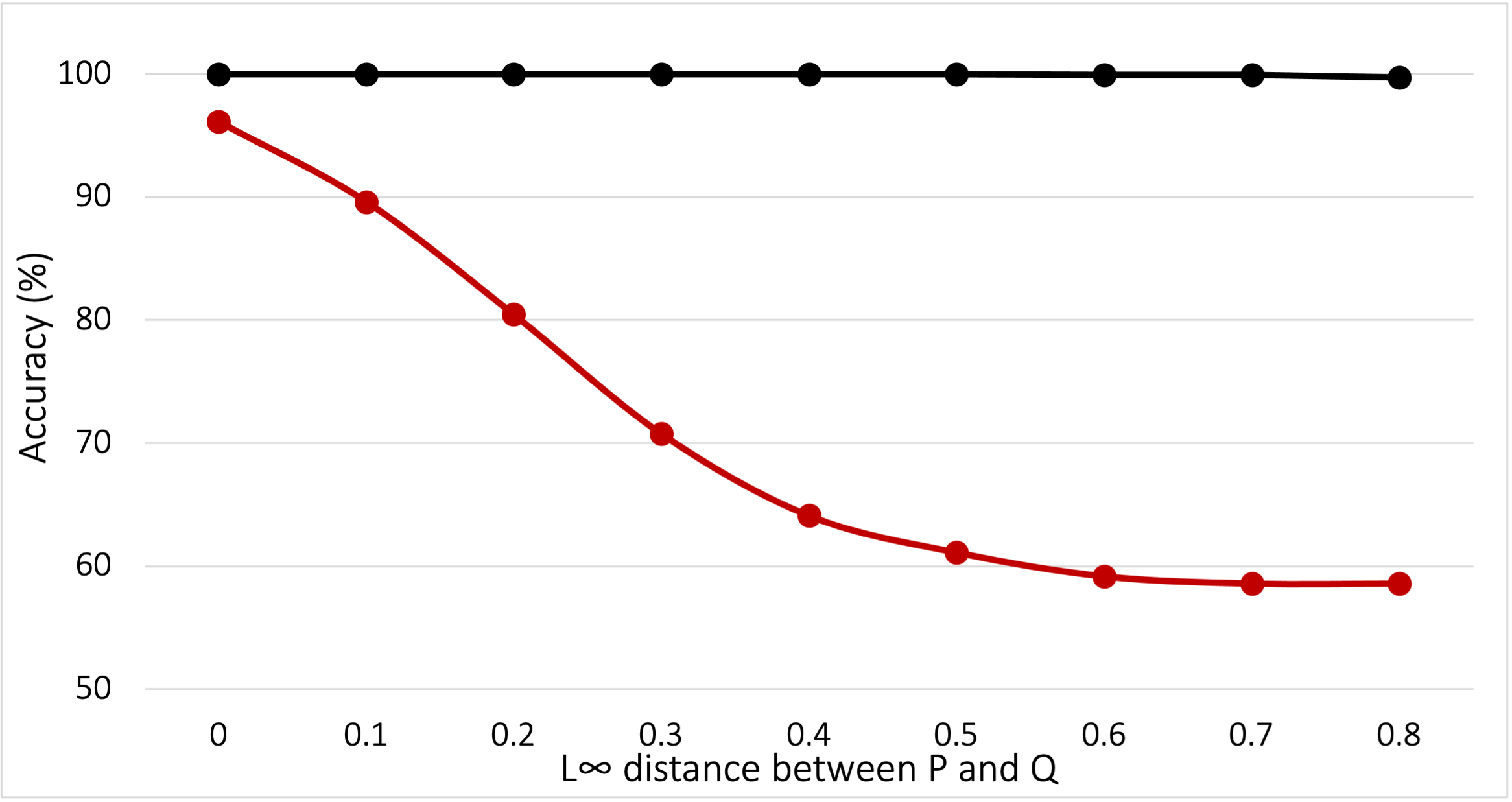}
\caption{The \emph{modulo-k task} is to classify whether the number of zeros in $x\in\{0,1\}^*$ is divisible by $k$ ($y=1$) or not $(y=0)$. We show the edge Markov Chains for generating examples for $\mathcal{L}_{mod-3}$, and is similar for $\mathcal{L}_{mod-4}$ and $\mathcal{L}_{mod-5}$ with additional states in between. The shifts are introduced similar to the parity language by perturbing the loop probabilities. We plot test accuracies for $\mathcal{L}_{mod-3},\mathcal{L}_{mod-4},\mathcal{L}_{mod-5}$ (in order from L-R) as a function of $\|P-Q\|_\infty$. For the baseline model (solid red line), the test accuracy drops significantly as $\|P-Q\|_\infty$ increases while our proposed SSAS approach (solid black line) ensures robustness to distribution shift.}
\label{fig:boundsk}
\end{figure}


\end{document}